\documentclass[conference]{IEEEtran}

\usepackage{balance}

\IEEEoverridecommandlockouts 
\usepackage[dvips]{graphicx}
\usepackage{listings}
\usepackage[OT4,T1]{fontenc}
\usepackage[cmex10]{amsmath}
\usepackage{url}
\usepackage{multirow}

\usepackage{algorithm}
\usepackage{mathtools}
\usepackage{multirow}
\usepackage{makecell}
\usepackage{algpseudocode}

\usepackage[utf8]{inputenc}

\usepackage{booktabs} 
\usepackage{array} 
\usepackage{paralist} 
\usepackage{verbatim} 
\usepackage{subfig} 

\usepackage{booktabs}
\usepackage{colortbl}

\title{Convolutional neural network compression for natural language processing}
\author{\IEEEauthorblockN{Krzysztof Wróbel\IEEEauthorrefmark{1},
Marcin Pietroń\IEEEauthorrefmark{1},
Maciej Wielgosz\IEEEauthorrefmark{1}, 
Michał Karwatowski\IEEEauthorrefmark{1} and
Kazimierz Wiatr\IEEEauthorrefmark{1}}
\IEEEauthorblockA{\IEEEauthorrefmark{1}AGH University of Science and Technology
Mickiewicza Av. 30, Krakow, Poland\\ Email: \{kwrobel,pietron,wielgosz,mkarwat,wiatr\}@agh.edu.pl}}

\begin{document}
\maketitle              

\begin{abstract}
Convolutional neural networks are modern models that are very efficient in 
many classification tasks. They were originally created for image processing purposes. Then some trials were performed to use them in different domains like natural language processing. The artificial intelligence systems (like humanoid robots) are very often based on embedded systems with constraints on memory, power consumption etc. Therefore convolutional neural network because of its memory capacity should be reduced to be mapped to given hardware.    
In this paper, results are presented of compressing the efficient convolutional neural networks for sentiment analysis. The main steps are quantization and pruning processes. The method responsible for mapping compressed network to FPGA and results of this implementation are presented. The described simulations showed that 5-bit width is enough to have no drop in accuracy from floating point version of the network. Additionally, significant memory footprint reduction was achieved (from 85\% up to 93\%).

\end{abstract}

\section{Introduction}
\IEEEoverridecommandlockouts\IEEEPARstart{N}{atural} Language processing (NLP) is considered one of three main pillars of Deep Learning along with image and video processing. Mobile devices are becoming increasingly dominant both in terms of their number as well as a computing load and network traffic. They also become more interactive in terms of NLP algorithms implemented for applications such as translation or voice-typing. Consequently, there is a significant benefit from reducing memory footprint and computational burden for deployment of NLP modules on embedded devices. It is worth noting that despite an abundance of research efforts in Deep Learning architectures compression for image processing \cite{1710.09282, 1803.03289, wrobel2018icaart}, there are just a few projects aiming at NLP neural architectures compression \cite{Raphael2017Compressing,Juan2017From, 1708.05963,kaul2018icaart}.

This paper shows a case study of employing common compression techniques for compressing NLP architectures. A series of quantization and pruning techniques were implemented and deployed in FPGA platform. The main goal of the paper was to examine feasibility and efficiency of using FPGAs in a domain of embedded neural computations. A set of common datasets were used to conduct reliable experiments. The experiments reviled a strong relationship between a size and the structure of datasets and performance of quantization and pruning methods used. All the layers of the neural architecture \cite{kim2014convolutional} were analyzed separately with respect to their ability to be quantized and pruned.

The structure of the paper is structured as follows. Section \ref{section:cnn} provides an overview of CNN architectures, quantization and pruning processes.  Section \ref{section:datasets} covers the dataset used for the experiments. Section \ref{section:network_architecture} describes neural model used for the experiments as well as FPGA implementation details of the architecture. Finally, section \ref{section:experiments} contains the results of the experiments. Section \ref{section:conclsions} summarizes contribution of this work.

\section{Convolutional Neural Network}
\label{section:cnn}
CNNs are composed of neurons that have learnable weights and biases. Each neuron receives inputs structured as multi-dimensional vectors (tensor), perform a convolution operation on this input it with multi-dimensional 
filters and then optionally follows outputs with pooling and non-linearity functions.  Typically, the layers of a CNN have neurons that generate outputs feature maps $y_i$, $i=1...N$ as follows 
\begin{equation}
\label{conv}
y_i = b_i + \sum_{j=1}^{M} F_{ij} * x_j
\end{equation}
where $F_{ij}$ are two-dimensional (2D) convolutional kernels of dimensions $H\times W$, $*$ represents the convolution operation and $b_i$ are the bias terms.   

The number of multiply-accumulate (MAC) operations and cycles spent on the execution of  in a practical implementation is often used as the metric for a complexity of a CNN. 
Assuming each output feature map $y_i$ has $P$ elements  (where P is equal to height multiplied by width of given feature map) the total number of MAC calculations for a convolutional filtering operation is ${MACs} = PHWMN$. Our role in quantization is to reduce the complexity of each one of these operations and in network compression through pruning the goal is to reduce the total number of operations. 

\subsection{Quantization process}

\begin{figure}[h]
\centering
\includegraphics[scale=0.4]{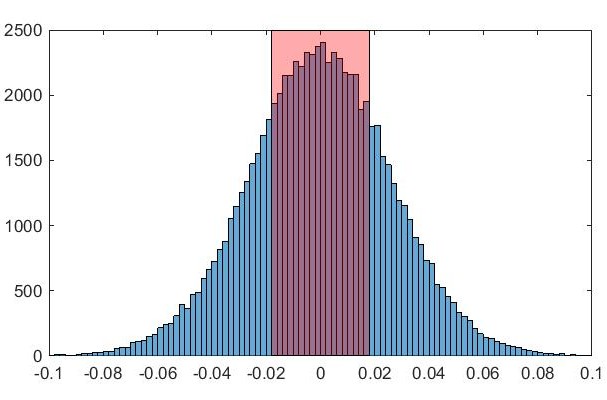}
\caption{Histogram of first convolutional layer weights with marked 50\% of weights cut out through pruning.}
\label{fig:histogram}
\end{figure}

\begin{algorithm*}[!h]
\caption{1D convolution layer with pruning\label{alg:prun_alg}}
\begin{algorithmic}[1]
\For{$outS$ in $output\_size$}
\For{$outFM$ in $output\_feature\_maps$}
\For{$inFM$ in $input\_feature\_maps$}
\For{$kerS$ in $kernel\_size$}
\If{$abs(weight[inFM][outFM][kerS])>=pruning\_threshold$}
\State $output(outS, outFM) += input(outS + kerS, inFM) * weight[inFM][outFM][kerS]$
\EndIf
\EndFor
\EndFor
\State $output(outH, outW) += bias[outW]$
\EndFor
\EndFor
\end{algorithmic}
\end{algorithm*}

Quantization is the procedure of constraining values from a continuous set or more dense domain to a relatively discrete set (e.g. integers), more sparse domain in which quantized input values will be represented. In our case, the domain is floating point representation. A floating-point number can be represented as

\begin{equation}
\label{fp}
\bf{ flp =  m \cdot b^e}
\end{equation}
$\bf{m} \in \mathcal{Z}$ is the mantissa, $\bf{b}=2$ is the base, and $\bf{e} \in \mathcal{Z}$ is the exponent. In case of single precision floating point format according to the 
IEEE-754 standard the mantissa is assigned 23bits, the exponent is assigned 8bits, and 1bit is assigned for a sign indicator. Therefore the set of values that can be defined 
by this format is described by 

\begin{equation}
	\text{flp}_{single}: \{ \pm 2^{-126}, \ldots , \pm (2 - 2^{-23}) \times 2^{127} \}.
\label{eq:ieee_754_single}
\end{equation}
Similarly, a reduced precision IEEE-754 mini-float format assigns 10bits of mantissa, 5bits of exponent and a sign bit.

Currently, GPUs with parallel processing being applied to machine learning operate mainly using the single precision format.  In contrast, embedded DSPs and the latest GPUs operate with 8-bit integer and 16-bit fixed-point processing that restricts numbers to a range 
\begin{equation}
	\text{fxp} : 2^{-{\bf frac\_bits}}\cdot\{-2^{{\bf total\_bits}-1},\ldots,2^{{\bf total\_bits}-1} - 1\}
\end{equation}
where ${\bf total\_bits}=8,16$ are conventional bit-widths and ${\bf frac\_bits}$ is a shift down (or up) that determines fractional length, or number of fractional bits, as well
as the integer length ${\bf int\_bits}={\bf total\_bits}-{\bf frac\_bits}-1$ for signed numerical representation.  When the fractional length varies over individual coefficients or 
data samples. This format is also referred to as {\em dynamic fixed-point} \cite{gysel2016ristretto}.

It is possible to define a general mapping from a set of floating-point data $x\in {\mathcal S}$ to fixed-point $q$ as follows (assuming signed representation)
\begin{equation}
	\label{eq:quant}
	q_{\text{fxp}} = {\mathcal Q}(x_{\text{flp}}) = \mu + \sigma \cdot \text{round}(\sigma^{-1}\cdot (x-\mu)). 
\end{equation}
%
%
%
In our case $\mu=0$ and $\sigma = 2^{-{\bf frac\_bits}}$ where 
\begin{equation}
	{\bf int\_bits}=\text{ceil}(\log_{2}(\max_{x\in{\mathcal S}} |x|)) 
\end{equation}
and ${\bf frac\_bits}={\bf total\_bits}-{\bf int\_bits}-1$.
The scaling factor $\sigma$ is essentially just a shift up or down.  A drawback is that a great deal of precision may be lost
if the distribution of the dataset ${\mathcal S}$ is skewed by a large mean.

Yet another approach can define the number of integer and fractional bits to represent regions of a distribution that will represent a large percentage of the range.  
In these cases, there will be saturation of a small percentage of the data, such as outliers, through the quantization procedure which may or may not affect the accuracy in a significant way. 
To determine the effects of saturation on can experiment with different saturation levels. Therefore histogram analysis is used to analyze outliers and set best levels of saturation.

Another approach is quantization that maps floating point values to integers:

\begin{equation}
	\label{eq:quant2}
	q_{\text{int}} = {\mathcal Q}(x_{\text{flp}}) = ceil((x -\mu) / ((max(X)-min(X)) \cdot \sigma^{-1}))
\end{equation}

The $\mu$ parameter can be set to $min(X)$ ($X$ is a input set of values to be quantized) or can be zero value. In a first case it is known as a asymmetric integer quantization (e.g. used in Tensorflow framework), in the second it is called symmetric. The compression system presented in the paper are based on the symmetric quantization and dynamic fixed point was also implemented.

All presented approaches are examples of linear quantization. It is possible also to use nonlinear version to minimize quantization loss but its hardware implementation is more sophisticated and it is more difficult to achieve gain in used hardware resources.

\begin{table}[h]
\caption{Statistics of 7 datasets: maximal number of words in a document, size of vocabulary, number of classes and number of documents. Test column indicates if cross-validation was performed or standard train-test split.}
\centering
\begin{tabular}{|@{\vrule width0ptheight9pt\enspace}l|r|r|r|r|r|}\hline
    & \textbf{Max length} & \textbf{Vocabulary} & \textbf{Classes} & \textbf{Documents} & \textbf{Test}\\\hline
    MR & 64 & 18767 & 2 & 10662 & CV\\\hline
    SST-1 & 61 & 17838 & 5 & 11855 & \\\hline
    SST-2 & 61 & 16190 & 2 & 9613 & \\\hline
    Subj & 128 & 21324 & 2 & 10000 & CV \\\hline
    TREC & 45 & 8766 & 6 & 5952 & \\\hline
    CR & 113 & 5341 & 2 & 3775 & CV \\\hline
    MPQA & 44 & 6248 & 2 & 10606 & CV \\\hline
    \end{tabular}
  \label{datasets_stats}
\end{table}

\subsection{Pruning}

After training neural model we acquire a set of weights for each trainable layer.
Those weights are not evenly distributed over the range of possible values for a selected data format. As presented in figure \ref{fig:histogram}, most weights are concentrated around 0 and are very close to it. Therefore, their impact on resulting activation value is not significant. The mechanism of pruning removes weights which values are below a certain threshold level. Authors of \cite{han2015learning} examined how pruning affects convolutional neural networks for image classification, showing little accuracy drop and high compression ratio.
In this work pruning is applied to 1D convolution layer according to algorithm \ref{alg:prun_alg}, as weights are concentrated around 0, for negative values threshold was calculated for an absolute value. Example from figure \ref{fig:histogram} presents that threshold level of 0.02 is sufficient to remove around 50\% of weights. Depending on specific network implementation storing weights may require significant amounts of memory, removing weights through pruning have a direct impact on lowering storage requirements.

The order in which quantization and pruning are applied should have little effect, especially for higher precisions. If quantization is applied first then weights from a certain range will be brought to the same value and pruning will only be able to remove none or all of them. If pruning will be applied first it will be possible to cut out weights more precisely. This will only have an effect on border quantization buckets and therefore will have any significant impact only for very small precisions.

\begin{table*}[!h]
\centering
\caption{Accuracy for SST2 dataset for 10 values of precision for each place. Suffix ``\_a'' means activations of the layer.}
\label{tab:SST2}
\begin{tabular}{|@{\vrule width0ptheight9pt\enspace}l|l|l|l|l|l|l|l|l|l|l|}\hline
Precision    & 32    & 16    & 8     & 7     & 6     & 5     & 4     & 3     & 2     & 1     \\\hline
embedding\_1 & 85.01 & 85.01 & 85.12 & 85.17 & 85.34 & 85.12 & 85.61 & 84.68 & 84.29 & 84.18 \\\hline
conv1d\_1    & 85.01 & 85.01 & 85.01 & 85.01 & 84.9  & 85.06 & 85.23 & 85.28 & 85.01 & 84.95 \\\hline
conv1d\_1\_a & 85.01 & 85.01 & 85.01 & 85.06 & 85.12 & 85.23 & 85.28 & 85.45 & 85.17 & 84.07 \\\hline
conv1d\_2    & 85.01 & 85.01 & 84.95 & 85.01 & 85.06 & 85.01 & 85.17 & 84.9  & 85.34 & 85.12 \\\hline
conv1d\_2\_a & 85.01 & 85.01 & 84.95 & 84.95 & 84.95 & 85.01 & 85.01 & 85.28 & 84.84 & 82.7  \\\hline
dense\_1     & 85.01 & 85.01 & 85.01 & 84.95 & 84.9  & 84.9  & 85.01 & 85.01 & 85.01 & 85.17 \\\hline
dense\_1\_a  & 85.01 & 85.01 & 85.06 & 85.06 & 85.12 & 84.9  & 84.84 & 83.42 & 69.91 & 52.33 \\\hline
dense\_2     & 85.01 & 85.01 & 84.95 & 85.01 & 85.06 & 85.06 & 84.95 & 85.06 & 85.17 & 84.9  \\\hline
all places   & 85.01 & 85.01 & 85.17 & 85.17 & 85.34 & 85.17 & 85.23 & 83.36 & 82.92 & 62.71\\\hline
\end{tabular}
\end{table*}

\section{Datasets}
\label{section:datasets}

Experiments are performed on the same datasets as used in \cite{kim2014convolutional}. Summary statistics of the datasets are provided in table \ref{datasets_stats}. Some datasets are divided into training, validation and testing data. If validation data is not specified then random 10\% of training is used for it. If testing data is not prepared then 10-fold cross validation is performed.

\begin{figure}[h]
\centering
\includegraphics[scale=0.2]{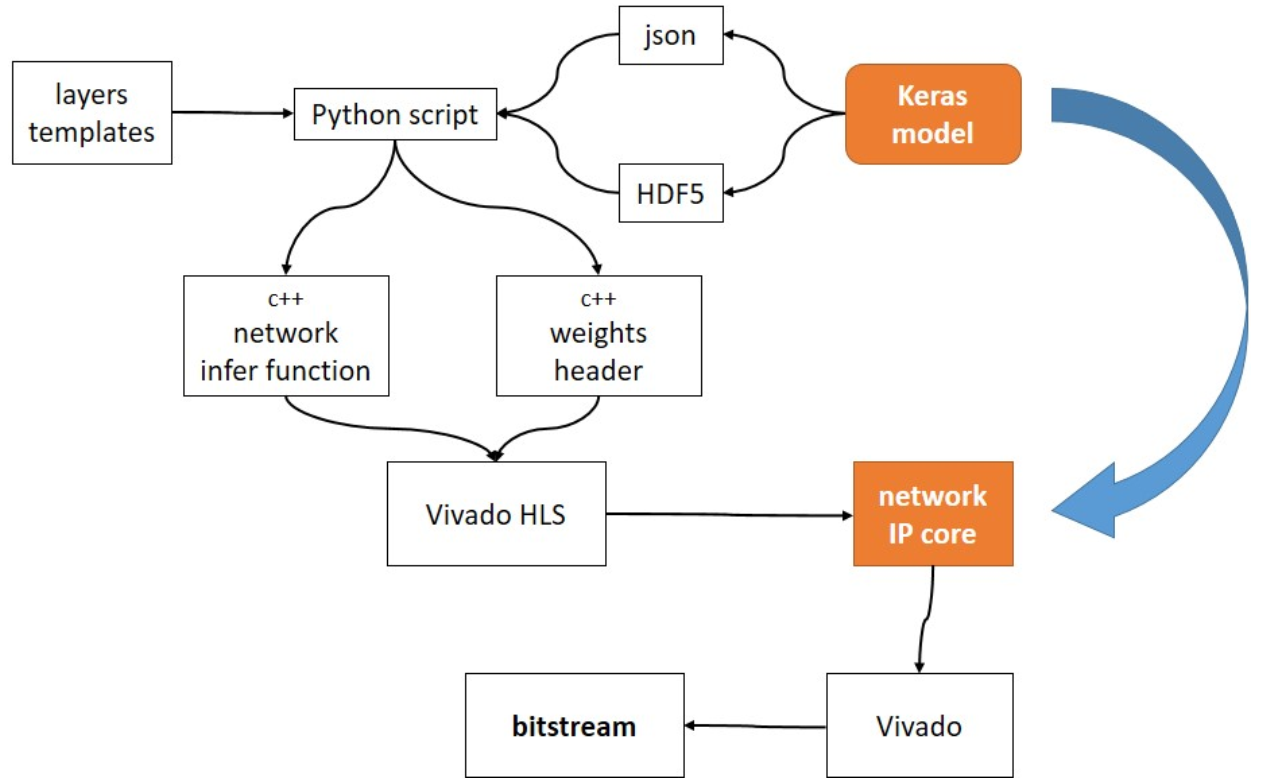}
\caption{Processing Keras model to FPGA IP core.}
\label{fig:schematic}
\end{figure}

\section{Network architecture}
\label{section:network_architecture}
We employed a model \textbf{CNN-static} from \cite{kim2014convolutional}. The model uses pre-trained 300-dimensional word embeddings GloVe \cite{pennington2014glove}, which are not fine-tuned during training. Unknown words are randomly initialized. Outputs of two parallel CNN layers with windows size 2 and 3 with 128 filters each are concatenated. The concatenation is connected to dense layer with 128 outputs. CNNs and dense layer have rectified linear units as activation functions. The last layer uses softmax activation function in case of many outputs and sigmoid activation function in case of binary classification. After each layer dropout was applied with probability 50\%. Training data was divided into mini-batches of size 128. The training involves Nadam \cite{Dozat2015IncorporatingNM} as optimizing algorithm. Training lasts 25 epochs, the best model is saved.

\subsection{FPGA Implementation}

Pruning mechanism is especially effective in FPGA implementations. On standard CPUs processing throughput improvement is not significant, multiply and accumulate operation is replaced by conditional, and on most modern processors both of those operations are supported by hardware. Greater improvements can be achieved when pruning is extensive enough to apply algorithms for sparse operations, such extreme pruning, however, will likely has a considerable impact on accuracy. When creating FPGA implementation of a certain neural network there is a higher degree of freedom. Its possible to hard code weights into logic during synthesis, as a result a complete cost of resolving condition if certain calculation should be performed, is moved from inference to compilation. Weights that will be removed by pruning are not included in the synthesized design, saving both logic and routing resources.

FPGA implementation can also give additional freedom for quantization adjustments, as it is not bound to any standard data width or format. On a standard CPU moving from 16 bit precision to 15 bit will not give any performance improvements, a calculation will be executed using the same hardware. For FPGAs however, it will result in synthesizing a smaller logic design with fewer connections, which can be beneficial for latency, throughput or calculations density.

\begin{table*}[!h]
\centering
\caption{Accuracy for MR dataset for 10 values of precision for each place.}
\label{tab:mr}
\begin{tabular}{|@{\vrule width0ptheight9pt\enspace}l|l|l|l|l|l|l|l|l|l|l|}\hline
Precision    & 32    & 16    & 8     & 7     & 6     & 5     & 4     & 3     & 2     & 1     \\\hline
embedding\_1 & 80.35 & 80.35 & 80.28 & 80.31 & 80.36 & 80.39 & 79.94 & 79.28 & 78.77 & 78.78 \\\hline
conv1d\_1    & 80.35 & 80.35 & 80.37 & 80.37 & 80.36 & 80.33 & 80.31 & 80.4  & 80.35 & 80.13 \\\hline
conv1d\_1\_a & 80.35 & 80.35 & 80.35 & 80.32 & 80.29 & 80.3  & 80.31 & 80.24 & 80.05 & 77.49 \\\hline
conv1d\_2    & 80.35 & 80.35 & 80.36 & 80.37 & 80.34 & 80.34 & 80.38 & 80.3  & 80.23 & 79.9  \\\hline
conv1d\_2\_a & 80.35 & 80.35 & 80.35 & 80.29 & 80.28 & 80.29 & 80.31 & 80.32 & 80.04 & 77.8  \\\hline
dense\_1     & 80.35 & 80.35 & 80.36 & 80.37 & 80.37 & 80.29 & 80.3  & 80.36 & 80.32 & 80.44 \\\hline
dense\_1\_a  & 80.35 & 80.35 & 80.43 & 80.41 & 80.31 & 80.07 & 79.33 & 75.92 & 64.95 & 52.86 \\\hline
dense\_2     & 80.35 & 80.35 & 80.36 & 80.38 & 80.37 & 80.33 & 80.32 & 80.37 & 80.38 & 80.2  \\\hline
all places   & 80.35 & 80.35 & 80.39 & 80.4  & 80.26 & 79.94 & 78.84 & 77.13 & 76.63 & 58.69 \\\hline
\end{tabular}
\end{table*}

Weights in the used neural network are mostly concentrated in Convolution operations, around 99\% of total trained weights, with 1\% in fully connected layers, as embeddings layer is not implemented in FPGA it is excluded from these calculations. Therefore FPGA implementation is concentrated on reducing and accelerating convolutional layer. The design was implemented and synthesized using Xilinx Vivado HLS environment according to algorithm \ref{alg:prun_alg}. To allow for more effective pruning, all weights are hard coded into logic. Loops from lines 2, 3 and 4 are fully unrolled. Due to limitations of synthesis tools layer from the used neural network could not be used. Therefore experimental results presented in section \ref{sec:experiments:fpga} are obtained from a smaller version of a similar network, the results, however, can be extrapolated. The process of transforming Keras model to an IP core that can be used in FPGA is presented in the figure \ref{fig:schematic}.
%
%

In the first step, Keras network model is created and trained, when accuracy is satisfactory network model and weights need to be exported. Weights from an HDF5 file are processed by a python script to C++ header format. It is important that their values will be known during compilation as Vivado HLS will not be able to create hard coded network otherwise. A separate header is created for each layer. Json file is used to create infer function of a neural network. This is a top level function that performs all the computations on input data. At this, we can decide how the data and weights should be represented. To validate the generation process a simulation was performed. If results are satisfactory, Vivado HLS will synthesize networks IP core. Next, the IP core is put into Vivado design and bitstream is generated.
%
%
%
%
%
%

\section{Precision reduction}
The precision reduction was applied to two types of places: weights of layers and outputs of activation functions. The process is performed by finding minimum and maximum of values to be reduced and uniformly dividing value range into $2^{bits}$ buckets. Each bucket has assigned a middle point of represented subrange. In case of layer weights, it is straightforward. Precision reduction of activation functions requires to run computation on training data and remember minimum and maximum of activation functions for chosen layers. During testing, it is possible that computed values will be lower than minimum and larger than maximum --- they need to be clipped.

The model used in this work has 5 layers: embeddings, two CNNs, and two dense layers. Activation functions are reduced after the two CNNs and the first dense layer. A number of bits used in each of the 8 places may be different. 

\begin{algorithm*}[h]
\caption{Random-restart hill climbing algorithm for searching maximal compression scheme.}\label{alg:hill}
\begin{algorithmic}
\ForAll{$places$}
\State $precisions[place]\gets 32$\Comment{default precision of 32 bits}
\EndFor
\State $original\_accuracy\gets \Call{test}{precisions}$
\State \Call{shuffle}{places}\Comment{shuffle order of places}
\Statex
\While{$precisions$ is changed}
\ForAll{$places$}
\State $n\gets precisions[place]$
\For{$precision\gets 1, n$}
\State $precisions[place]\gets precision$
\State $accuracy\gets \Call{test}{precisions}$
\If{$accuracy\ge original\_accuracy \cdot threshold$}
\State \textbf{break}
\EndIf
\EndFor
\EndFor
\EndWhile
\end{algorithmic}
\end{algorithm*}

\begin{table}[h]
  \caption{Model size reduction of the 50 runs of the algorithm \ref{alg:hill}}.
\centering
\begin{tabular}{|@{\vrule width0ptheight9pt\enspace}l|r|r|r|}\hline
    &  Size reduction & Size [MB] & Size of embeddings\\\hline
    MR &  84.69\% & 3.42 & 98.14\% \\\hline
    SST-1 &  87.52\% & 2.66 & 96.12\% \\\hline
    SST-2 &  93.76\% & 1.21 & 95.66\% \\\hline
    Subj &  87.73\% & 3.10 & 98.43\% \\\hline
    TREC &  88.24\% & 1.28 & 97.89\% \\\hline
    CR &  85.33\% & 1.02 & 93.39\% \\\hline
    MPQA & 85.33\% & 1.18 & 95.09\% \\\hline
    \end{tabular}
  \label{tab:model_compression}
\end{table}

\begin{table*}[h]
  \caption{The best precision settings found by algorithm \ref{alg:hill} for each dataset. Accuracy and size of the model is also reported.}
\centering
\begin{tabular}{|@{\vrule width0ptheight9pt\enspace}l|r|r|r|r|r|r|r|}\hline
    & MR & SST-1 & SST-2 & Subj & TREC & CR & MPQA \\\hline
    embedding\_1 & 5 & 4 & 2 & 4 & 4 & 5 & 5 \\\hline
    conv1d\_1 & 2 & 5 & 1 & 1 & 1 & 2 & 2 \\\hline
    conv1d\_1\_a & 5 & 4 & 3 & 2 & 2 & 5 & 7 \\\hline
    conv1d\_2 & 3 & 3 & 2 & 2 & 1 & 3 & 2 \\\hline
    conv1d\_2\_a & 4 & 6 & 2 & 3 & 3 & 3 & 6 \\\hline
    dense\_1 & 1 & 4 & 4 & 3 & 1 & 2 & 3 \\\hline
    dense\_1\_a & 7 & 8 & 3 & 6 & 5 & 6 & 6 \\\hline
    dense\_2 & 2 & 1 & 3 & 2 & 3 & 2 & 1 \\\hline
    Accuracy & 80.21\% & 46.92\% & 84.57\% & 92.75\% & 90.40\% & 83.15\% & 88.91\% \\\hline
    Size [MB] & 3.4194 & 2.655 & 1.210 & 3.099 & 1.281 & 1.023 & 1.175 \\\hline
    \end{tabular}
  \label{best_settings}
\end{table*}

\section{Experiments}
\label{section:experiments}

\begin{figure*}[h]
\centering
\includegraphics[scale=0.3]{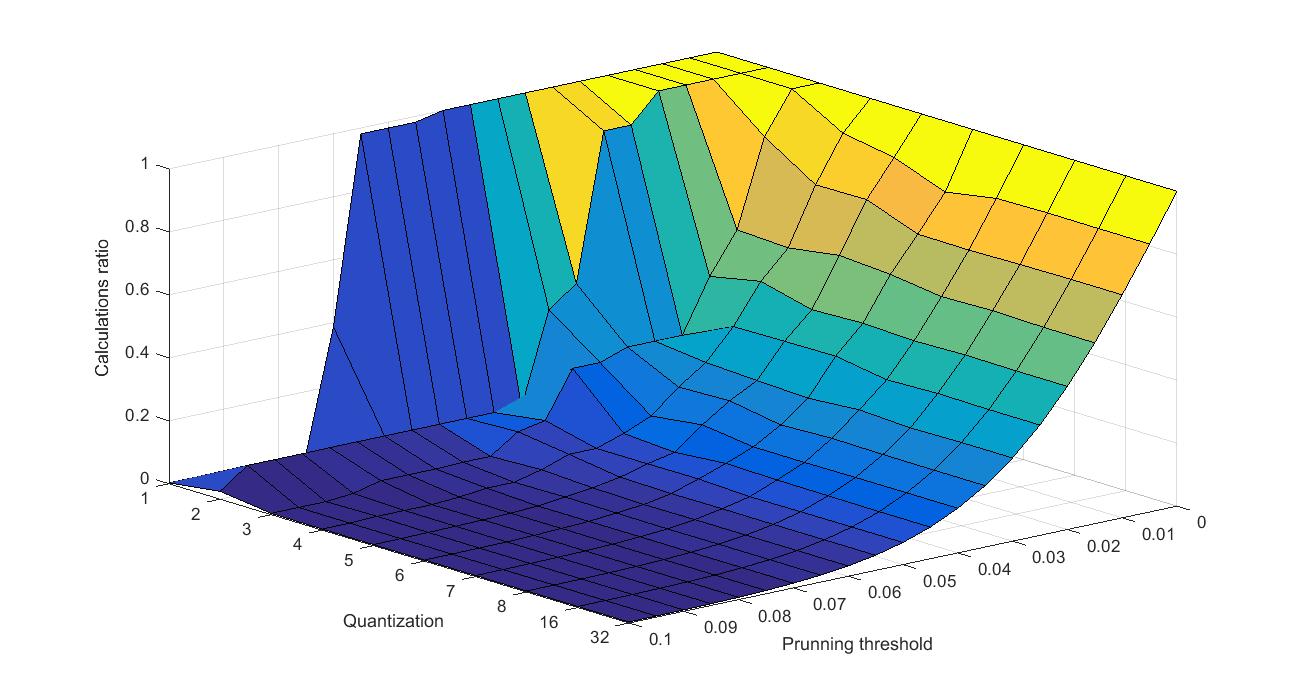}
\caption{Ratio of calculations performed after pruning to number of calculations that would be performed without pruning.}
\label{fig:pruning}
\end{figure*}

\begin{figure*}[h]
\centering
\includegraphics[scale=0.3]{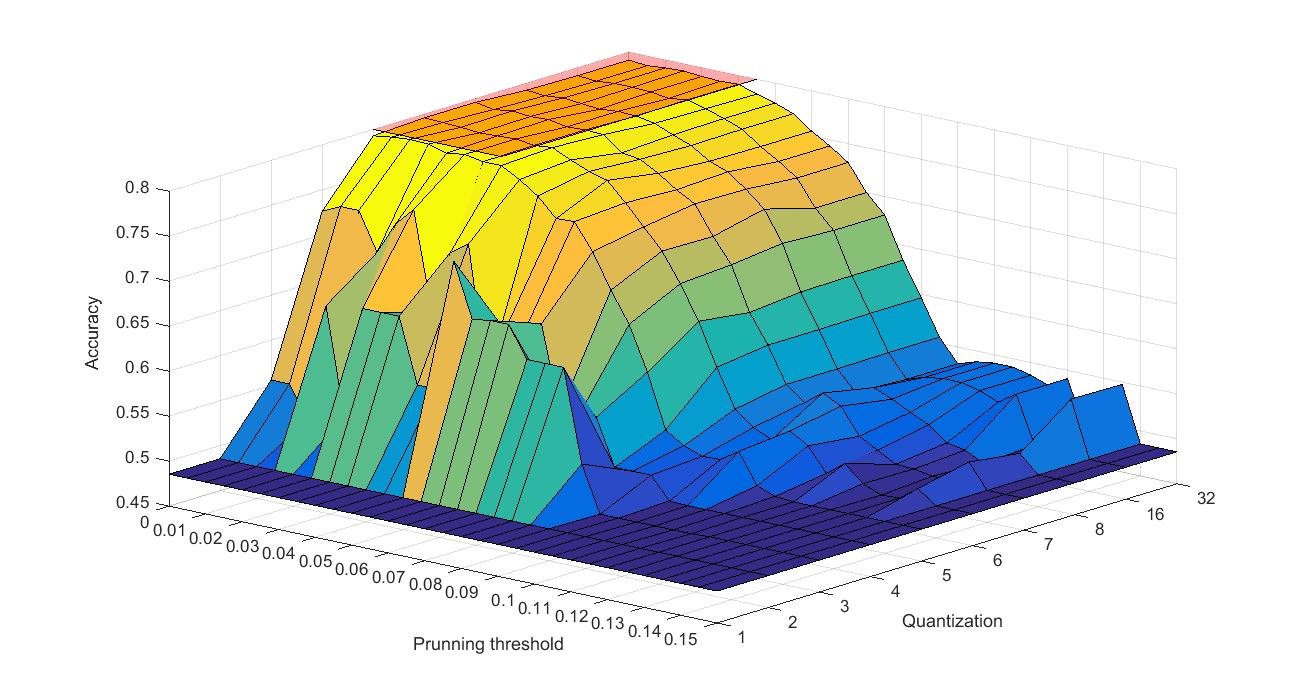}
\caption{Accuracy results while network reduction through quantization and pruning.}
\label{fig:accuracy}
\end{figure*}

The first set of experiments involved precision reduction to the same number of bits on every combination of the 8 places. Tested number of bits includes 32, 16, 8, 7, 6, 5, 4, 3, 2 and 1. It gives $2^8 \cdot 10$ experiments for each dataset/fold. All presented results are based on integer quantization. In all experiments this type of quantization gives little bit better accuracy than in dynamic fixed point (difference up to 2\%).

The results show that precision reduction to 16 bits does not change accuracy. Embeddings can be reduced to 5 bits, a majority of datasets weights of convolutional layers can be reduced to 4 bits. It is important to mark that convolutional layers are parallel and reducing one of the layers to 1 bit does not change overall accuracy significantly because the second is operating with full precision.

Tables \ref{tab:SST2} and \ref{tab:mr} presents results for SST-2 and MR datasets for each place precision reduced separately. In the case of 32 bits there is no reduction as it is default precision. 16 bits do not change accuracy results. Accuracy increases for some settings. The maximal drop in accuracy less than 1\% is observed down to 4--5 bits. Usually, quantization of activations decreases accuracy more than quantizations of weights. Precision reduction to less than 4 bits of activations of first dense layer affects negatively accuracy the most. The tables present also results for precision reduction of all places. a joint reduction can be better than only for one place.

The second experiment was focused on finding the strongest precision reduction with arbitrary chosen maximal accuracy decrease. Here we are not limiting the model reduction to the same number of bits for each place. In order to test every combination including 10 values of a number of bits, it would be necessary to perform $10^8$ tests, which is not computable in a reasonable time.
To tackle this problem a random-restart hill climbing algorithm was employed. The algorithm starts with a randomly chosen order of places to be reduced. Then iterates the places in a cyclic way until does not increase model compression. For each layer, it tries to find the smallest number of bits that does not decrease accuracy more than a set threshold. The searching method using algorithm \ref{alg:hill} was restarted 50 times. In this work, we set maximal decrease as 0.2\% of accuracy without precision reduction.

Table \ref{tab:model_compression} shows results of precision reduction for each dataset. In comparison to 32 bit representation, the size of models can be reduced by more than 84\%. However, the most space is taken by word embeddings (more than 93\%). It could be further reduced by limiting vocabulary and reducing dimensions of embeddings. Another approach was to order the places by their influence on model size, in this case starting from optimizing embedding layer. The approach has not achieved better results than the algorithm \ref{alg:hill}. The same precision bits parameters were obtained for two datasets.

The best found precision settings are shown in table \ref{best_settings}. All precision values are equal to or less than 8 bits.

The third experiment involved a combination of quantization and pruning. It was performed on MR database, the number of bits changed in the same way as in previous experiments. Pruning was applied according to algorithm \ref{alg:prun_alg}. pruning threshold started from 0, i.e. no pruning, with 0.005 step to 0.15, where all weights were removed and no multiply and accumulate operations were performed in convolutional layers.
Figure \ref{fig:pruning} presents extensiveness of pruning for each data precision. It is visible that the ratio quickly drops, it is a result of Gaussian distribution of weights concentrated around 0. When quantization steps are greater than pruning threshold step ratio decreases more gradually.

Pruning impact on the accuracy of the network is presented in figure \ref{fig:accuracy}. Its effect on accuracy is highly separate from quantization. A significant drop in accuracy appears below 5 bit precision, and over 0.03-0.035 pruning threshold. Between these values and no reduction point, a rectangular plateau of only slight variations in precision can be observed.

\subsection{FPGA results}\label{sec:experiments:fpga}

\begin{table}[h]
  \caption{FPGA resource utilization for selected quantization and pruning settings.}
\centering
\begin{tabular}{|@{\vrule width0ptheight9pt\enspace}c|c|c|}
\hline
          & \multicolumn{2}{c|}{pruning}                                                                                                                         \\ \hline
Precision & 0                                                                         & 0.035                                                                     \\ \hline
32        & \begin{tabular}[c]{@{}c@{}}FF 8118\\ LUT 60750\end{tabular} & \begin{tabular}[c]{@{}c@{}}FF 4449\\ LUT 14906\end{tabular} \\ \hline
5         & \begin{tabular}[c]{@{}c@{}}FF 2560\\ LUT 14957\end{tabular} & \begin{tabular}[c]{@{}c@{}}FF 2457\\ LUT 14382\end{tabular} \\ \hline
    \end{tabular}
  \label{tab:fpga_res}
\end{table}

Described quantization and pruning mechanisms were implemented in FPGA to examine their impact on logic utilization. Experiments were performed on a smaller convolutional layer, with an input of size 64 and 35 feature maps, the filter of size 2 and output of size 63 and 16 feature maps. We checked what effects gives pruning and quantization separately and in conjunction. Experiments were performed on Xilinx Virtex-7 FPGA VC707 Evaluation Kit with Virtex 7 series FPGA. Test points were selected so the accuracy remained within 1\% of the original. Results are summarized in table \ref{tab:fpga_res}. When both techniques are used flip flips utilization drops $3.3$ times and look up tables utilization drops $4.2$ times, while having a little effect on the accuracy.
%
%

\section{Conclusions and future work}
\label{section:conclsions}
This paper considers several compression and pruning techniques for convolutional neural networks in sentiment analysis. The results of the experiments show that the size of the model may be reduced more than 84\% with the little performance degradation of approximately 1\%. Word embeddings can be compressed even more reaching 93\%.   
Future work will concentrate on boosting the performance of given convolutional network by changing the architecture using deeper model and modifying filter scheme. The next issue will focus on using memetic algorithm and reinforcement techniques to find a configuration of the network which gives the best compression. 
\balance

\end{document}